\documentclass[12pt,a4paper, amsfonts, amssymb, amsmath, reprint, showkeys, nofootinbib, twoside,superscriptaddress,onecolumn]{revtex4-1}
\usepackage[utf8]{inputenc}
\usepackage[T1]{fontenc}
\renewcommand{\abstractname}{ABSTRACT}
\usepackage{xcolor}
\usepackage{caption}
\usepackage{graphicx}
\usepackage{threeparttable}
\usepackage{multirow}
\usepackage{hyperref}
\usepackage{wrapfig}
\usepackage{lineno}
\usepackage{enumerate}
\usepackage{booktabs}
\usepackage{amsmath}
\usepackage{caption}
\usepackage{bm}
\makeatletter
\renewcommand\subsubsection{\@startsection{subsubsection}{3}{0pt}{1em}{-1em}{\normalfont\bfseries\itshape}}
\makeatother
\begin{document}
	%\title{A Forensic Multi-Agent Framework Empowered by Large Language Models for Automated Cause-of-Death Determination in China}
	%\title{FEAT: A Forensic Multi-Agent AI System Empowered by Domain-Adapted Large Language Models for Autonomous Cause-of-Death Analyses}
	%\title{FEAT: A Multi-Agent Forensic AI System with Domain-Adapted LLMs for Automated Cause-of-Death Analysis}
	% 2025 0804 申忱 change
	\title{FEAT: A Multi-Agent Forensic AI System with Domain-Adapted Large Language Model for Automated Cause-of-Death Analysis}
	
	\affiliation{Key Laboratory of National Ministry of Health for Forensic Sciences, School of Medicine \& Forensics, Health Science Center, Xi'an Jiaotong University, Xi'an, Shaanxi 710049, China}
	\affiliation{School of Mathematics and Statistics, Xi'an Jiaotong University, Xi'an, Shaanxi 710149, China}
	\affiliation{Hebei Key Laboratory of Forensic Medicine, College of Forensic Medicine, Hebei Medical University, Shijiazhuang, Hebei 050017, China}
	\affiliation{Key Laboratory of Biomedical Information Engineering of Ministry of Education, School of Life Science and Technology, Xi'an Jiaotong University, Xi'an, Shaanxi 710049, China}
	\affiliation{Faculty of Forensic Medicine, Guangdong Province Translational Forensic Medicine Engineering Technology Research Center, Zhongshan School of Medicine, Sun Yat-Sen University, Guangzhou, Guangdong 510080, China} 
	\affiliation{Department of Forensic Medicine, Xinxiang Medical University, Xinxiang, Henan 453003, China}
	\affiliation{Department of Forensic Medicine, School of Basic Medical Sciences, Fudan University, Shanghai 200032, China}
	\affiliation{Department of Forensic Medicine, Tongji Medical College, Huazhong University of Science and Technology, Wuhan, Hubei 430030, China}
	\affiliation{Shanghai Key Laboratory of Forensic Medicine, Shanghai Forensic Service Platform, Academy of Forensic Science, Ministry of Justice, Shanghai 200063, China}
	\affiliation{School of Forensic Medicine of Jining Medical University, Jining, Shandong 272067, China}
	\affiliation{Shaanxi Baimei Forensic Appraisal Institutions, Xi'an, Shaanxi 710068, China}
	%\affiliation{These authors contributed equally to this work}
	%\affiliation{Corresponding Authors}
	
	\author{Chen Shen}
	\thanks{These authors contributed equally to this work}
\affiliation{Key Laboratory of National Ministry of Health for Forensic Sciences, School of Medicine \& Forensics, Health Science Center, Xi'an Jiaotong University, Xi'an, Shaanxi 710049, China}
	%\author{Fan Wang}
	%\affiliation{Key Laboratory of Biomedical Information Engineering of Ministry of Education, School of Life Science and Technology, Xi'an Jiaotong University, Xi'an, Shaanxi 710049, China}
	%\affiliation{These authors contributed equally to this work}	
	\author{Wanqing Zhang}
	\thanks{These authors contributed equally to this work}
\affiliation{Key Laboratory of National Ministry of Health for Forensic Sciences, School of Medicine \& Forensics, Health Science Center, Xi'an Jiaotong University, Xi'an, Shaanxi 710049, China}
	\author{Kehan Li}
	%\thanks{These authors contributed equally to this work}
	\affiliation{Key Laboratory of Biomedical Information Engineering of Ministry of Education, School of Life Science and Technology, Xi'an Jiaotong University, Xi'an, Shaanxi 710049, China}
	\author{Erwen Huang}
	\affiliation{Faculty of Forensic Medicine, Guangdong Province Translational Forensic Medicine Engineering Technology Research Center, Zhongshan School of Medicine, Sun Yat-Sen University, Guangzhou, Guangdong 510080, China}
	\author{Haitao Bi}
\affiliation{Hebei Key Laboratory of Forensic Medicine, College of Forensic Medicine, Hebei Medical University, Shijiazhuang, Hebei 050017, China}
	\author{Aiying Fan}
	\affiliation{Department of Forensic Medicine, Xinxiang Medical University, Xinxiang, Henan 453003, China}
	\author{Yiwen Shen}
	\affiliation{Department of Forensic Medicine, School of Basic Medical Sciences, Fudan University, Shanghai 200032, China}
	\author{Hongmei Dong}
	\affiliation{Department of Forensic Medicine, Tongji Medical College, Huazhong University of Science and Technology, Wuhan, Hubei 430030, China}
	\author{Ji Zhang}
	\affiliation{Shanghai Key Laboratory of Forensic Medicine, Shanghai Forensic Service Platform, Academy of Forensic Science, Ministry of Justice, Shanghai 200063, China}
	\author{Yuming Shao}
	\affiliation{School of Forensic Medicine of Jining Medical University, Jining, Shandong 272067, China}
	\author{Zengjia Liu}
	\affiliation{School of Forensic Medicine of Jining Medical University, Jining, Shandong 272067, China}
	\author{Xinshe Liu}
\affiliation{Key Laboratory of National Ministry of Health for Forensic Sciences, School of Medicine \& Forensics, Health Science Center, Xi'an Jiaotong University, Xi'an, Shaanxi 710049, China}
	\author{Tao Li}
\affiliation{Key Laboratory of National Ministry of Health for Forensic Sciences, School of Medicine \& Forensics, Health Science Center, Xi'an Jiaotong University, Xi'an, Shaanxi 710049, China}
	\author{Chunxia Yan}
\affiliation{Key Laboratory of National Ministry of Health for Forensic Sciences, School of Medicine \& Forensics, Health Science Center, Xi'an Jiaotong University, Xi'an, Shaanxi 710049, China}
	\author{Shuanliang Fan}
\affiliation{Key Laboratory of National Ministry of Health for Forensic Sciences, School of Medicine \& Forensics, Health Science Center, Xi'an Jiaotong University, Xi'an, Shaanxi 710049, China}
	\author{Di Wu}
	\affiliation{Shaanxi Baimei Forensic Appraisal Institutions, Xi'an, Shaanxi 710068, China}
	\author{Jianhua Ma}
	\thanks{Corresponding Authors (chunfeng.lian@xjtu.edu.cn; wzy218@xjtu.edu.cn; cong6406@hebmu.edu.cn; jhma@xjtu.edu.cn)}
	\affiliation{Key Laboratory of Biomedical Information Engineering of Ministry of Education, School of Life Science and Technology, Xi'an Jiaotong University, Xi'an, Shaanxi 710049, China}	
	\author{Bin Cong}
	\thanks{Corresponding Authors (chunfeng.lian@xjtu.edu.cn; wzy218@xjtu.edu.cn; cong6406@hebmu.edu.cn; jhma@xjtu.edu.cn)}
\affiliation{Hebei Key Laboratory of Forensic Medicine, College of Forensic Medicine, Hebei Medical University, Shijiazhuang, Hebei 050017, China}
	\author{Zhenyuan Wang}
	\thanks{Corresponding Authors (chunfeng.lian@xjtu.edu.cn; wzy218@xjtu.edu.cn; cong6406@hebmu.edu.cn; jhma@xjtu.edu.cn)}
\affiliation{Key Laboratory of National Ministry of Health for Forensic Sciences, School of Medicine \& Forensics, Health Science Center, Xi'an Jiaotong University, Xi'an, Shaanxi 710049, China}
	\author{Chunfeng Lian}
	\thanks{Corresponding Authors (chunfeng.lian@xjtu.edu.cn; wzy218@xjtu.edu.cn; cong6406@hebmu.edu.cn; jhma@xjtu.edu.cn)}
	%\email{chunfeng.lian@xjtu.edu.cn}
	\affiliation{School of Mathematics and Statistics, Xi'an Jiaotong University, Xi'an, Shaanxi 710149, China}
	%\affiliation{These authors contributed equally to this work}
	% 所有作者除了我和帮我画图的师妹还有可涵剩下都是法医老师（没有认识的，那个丛斌是院士，王老师说给通讯表示尊重）。。。

	\begin{abstract}
		
	\textbf{\abstractname}
	\vspace{1em}  
Forensic cause-of-death determination faces systemic challenges, including workforce shortages and diagnostic variability, particularly in high-volume systems like China's medicolegal infrastructure. 
We introduce FEAT (ForEnsic AgenT), a multi-agent AI framework that automates and standardizes death investigations through a domain-adapted large language model. FEAT's application-oriented architecture integrates: (i) a central Planner for task decomposition, (ii) specialized Local Solvers for evidence analysis, (iii) a Memory \& Reflection module for iterative refinement, and (iv) a Global Solver for conclusion synthesis. The system employs tool-augmented reasoning, hierarchical retrieval-augmented generation, forensic-tuned LLMs, and human-in-the-loop feedback to ensure legal and medical validity. In evaluations across diverse Chinese case cohorts, FEAT outperformed state-of-the-art AI systems in both long-form autopsy analyses and concise cause-of-death conclusions. 
It demonstrated robust generalization across six geographic regions and achieved high expert concordance in blinded validations. Senior pathologists validated FEAT's outputs as comparable to those of human experts, with improved detection of subtle evidentiary nuances.
To our knowledge, FEAT is the first LLM-based AI agent system dedicated to forensic medicine, offering scalable, consistent death certification while maintaining expert-level rigor. 
By integrating AI efficiency with human oversight, this work could advance equitable access to reliable medicolegal services while addressing critical capacity constraints in forensic systems.

	\end{abstract}
	
	\maketitle
	
\section*{Introduction} \label{sec:intro}

Forensic pathology serves a fundamental function in justice and public health, providing critical investigations into unnatural or unexplained deaths~\cite{vital_1,vital_2,vital_3,vital_4}. 
Globally, demand for forensic services has increased due to population growth, urbanization, and heightened legal and public health oversight~\cite{china_1,china_2}. 
In China, this trend is particularly pronounced, with tens of thousands of annual medicolegal cases (homicides, accidents, and unexplained hospital deaths) requiring expert analysis~\cite{china_3,china_4}. 
Yet the system is severely strained: roughly $12,000$ certified forensic pathologists serve a population exceeding $1.4$ billion, distributed across public security bureaus, procuratorates, and academic institutions~\cite{workforce_size,shortage}. 
Chronic workforce shortages force excessive caseloads, with many practitioners exceeding the recommended annual limit of $250$ autopsies---a threshold beyond which accuracy declines~\cite{num_autopsy_1,num_autopsy_2}. 
These backlogs delay legal resolutions and exacerbate distress for bereaved families awaiting answers~\cite{backlogs_1,backlogs_2}.

%%%%%%%%%%%%%%%%%%%
The challenge extends beyond caseload volume to the inherent complexity of cause-of-death analysis. 
Modern forensic investigations require synthesizing multifaceted evidence, including autopsy findings, toxicology reports, medical histories, crime scene data, and sometimes sociocultural context~\cite{information_1,information_2,information_3,information_4}. 
While seasoned experts develop nuanced interpretive skills over years (e.g., differentiating postmortem artifacts from antemortem trauma), case distribution often precludes exclusive involvement of senior specialists. 
China's regional disparities and decentralized system further exacerbate variability in forensic quality, with under-resourced areas more prone to oversights or inconsistent conclusions~\cite{diversity,wrong_1,wrong_2}. 
Even experienced practitioners routinely consult references or peers in complex cases, underscoring that death determination blends scientific rigor with expert judgment~\cite{pathologists}. 
These systemic pressures, i.e., rising caseloads, expertise shortages, and analytical complexity, highlight the need for computational solutions, particularly those leveraging artificial intelligence (AI)~\cite{ai-fp,ai-fp2}.

%%%%%%%%%%%%%%%%%%%%%%%%%Technological Opportunity
Recent advances in large language models (LLMs) and LLM-based agent AI systems offer transformative potential for automating reasoning in complex tasks like clinical and forensic medicine. 
State-of-the-art LLMs (e.g., GPT~\cite{gpt4}, Claude~\cite{claude}, and DeepSeek~\cite{deepseek-r1}) exhibit exceptional natural language understanding, medical knowledge recall, and multi-step reasoning capabilities---qualities critical for cause-of-death analyses\cite{gpt4_med}. 
The structured reasoning capabilities of these models can be substantially enhanced through Chain-of-Thought (CoT) prompting~\cite{cot}, which encourages step-by-step, transparent reasoning that closely aligns with clinical and forensic workflows.
Further progress comes from autonomous agent frameworks like Reasoning and Acting (ReAct)~\cite{react}, where LLMs dynamically interleave logical deduction with tool use (e.g., querying databases or performing calculations). 
More sophisticated approaches such as Tree-of-Thought (ToT)~\cite{tot} enable parallel exploration of competing hypotheses (e.g., accident vs. homicide vs. natural causes), while retrieval-augmented generation (RAG)~\cite{rag} grounds outputs in authoritative sources like medical literature or legal statutes. 
Together, these cutting-edge technologies could enable AI systems to parse autopsy reports, formulate investigative plans, consult domain-specific resources on demand, and converge on defensible cause-of-death conclusions, potentially leading to standardized, scalable forensic analyses that address current human resource limitations.

%%%%%%%%%%%%%%%%%%%%%%%%%Limitations of Prior Work
However, despite AI's success in clinical applications (e.g., diagnostic support~\cite{amie} and medical imaging~\cite{gigapath}), forensic medicine remains an underdeveloped frontier. 
Early forensic expert systems were narrowly task-specific (e.g., postmortem autolysis analysis~\cite{fpath} and tissue-level forensic diagnosis~\cite{songci}) and incapable of holistic case evaluation. 
Even recent medical AI models often produce single-pass, contextually shallow outputs lacking justification chains or subproblem decomposition~\cite{medfound,medpalm}. 
Moreover, most systems lack tool-use capabilities, increasing hallucination risks when external validation (e.g., textbook checks) is omitted~\cite{med_agent_r1,med_agent_r2}.
Emerging agent frameworks (e.g., MedAgent~\cite{medagent} and related efforts~\cite{Mdagents,Mmedagent}) demonstrate progress by integrating LLM planners with specialized modules for clinical tasks. 
However, these systems primarily target general healthcare (e.g., symptom-based diagnosis) and rely on proprietary, general-purpose LLMs (e.g., GPT-4O~\cite{gpt4o}, Claude~\cite{claude}) untrained on forensic corpora. 
This poses acute challenges for forensic contexts, where autopsy reports rely on domain-specific terminology and localized references. 
For instance, neither GPT-4O nor Claude has been exposed to sufficient Chinese forensic case data during training, which may lead to superficially plausible but factually unreliable outputs due to vocabulary gaps and ungrounded inferences. 
Consequently, there is an urgent need to develop agent AI systems explicitly tailored for autonomous cause-of-death analyses. Such systems must integrate forensic domain expertise, tool-augmented reasoning capabilities, and localization to align with Chinese medicolegal practices. Addressing this gap could substantially enhance forensic analyses, delivering standardized and scalable solutions that mitigate current limitations in human expertise and resource availability.

\begin{figure*}[htbp]
	\nolinenumbers
	\centering
	\captionsetup{font={tiny,bf,stretch=1.25},justification=raggedright}
	\includegraphics[width=\linewidth]{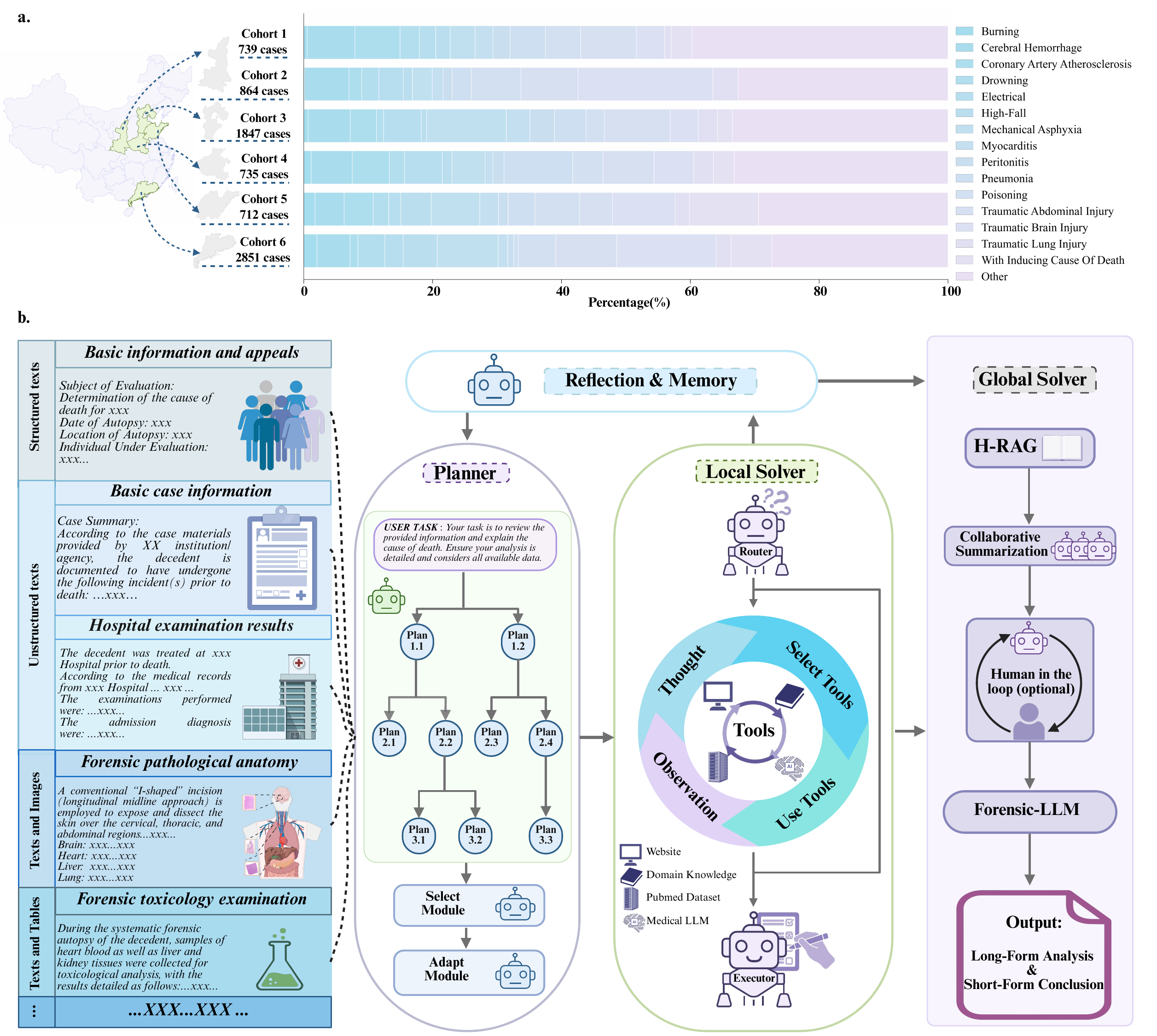}
	\caption{Overview of the FEAT system and data. (a) Category composition of the six evaluation cohorts (C1--C6). Stacked bars show the percentage distribution across cause-of-death categories (legend at right; cohort case counts labeled at left). (b) The system is: (1) FEAT ingests multi-source materials. (2) The Planner decomposes the case into a hierarchical plan and adapts it to the instance. (3) Local Solvers, coordinated by a Router, follow a ReAct cycle to generate evidence-grounded findings using resources such as a forensic textbook/vector store, PubMed retrieval, websites, and a medical LLM; an Executor synthesizes subtask results. (4) Reflection \& Memory maintains a dynamic case file, audits intermediate outputs for completeness and consistency, and triggers replanning when gaps or contradictions are detected---forming an iterative loop. (5) The Global Solver performs collaborative summarization and employs hierarchical retrieval-augmented generation to retrieve similar cases and authoritative references; with optional human-in-the-loop review, it invokes a Forensic-LLM to produce court-ready outputs: a long-form analysis and a short-form conclusion.}
	\label{fig:agent}
\end{figure*}

In this work, we present FEAT (ForEnsic AgenT), a multi-agent AI system specifically designed for autonomous forensic cause-of-death analysis and decision support. FEAT processes heterogeneous multi-source inputs (e.g., basic information, pathological anatomy, toxicology reports) through a coordinated ensemble of role-specific AI agents, mimicking the collaborative workflow of human forensic specialists. As illustrated in FIG.~\ref{fig:agent}, the system's modular architecture integrates: (i) a Planner for self-discovered task decomposition, (ii) Local Solvers employing tool-augmented ReAct reasoning, (iii) a Reflection \& Memory module for iterative self-correction, and (iv) a Global Solver combining hierarchical RAG (H-RAG) with locally fine-tuned LLMs---collectively featuring transparent, auditable medicolegal reasoning.
To ensure domain relevance, we curated and annotated the first comprehensive Chinese-language medicolegal corpus, facilitating both system adaptation and rigorous evaluation. Moreover, FEAT incorporates a human-in-the-loop interface allowing forensic experts to validate and refine outputs, ensuring compliance with professional standards.
Experimental results demonstrate FEAT's superiority over single-agent LLMs and existing medical AI agent systems (e.g., $3.2\%$ and $10.7\%$ accuracy improvements against the strongest baseline in terms of long-form analysis and short-form conclusion, respectively; statistically significant generalizability across geographically heterogeneous cohorts).
In blinded evaluations, senior forensic pathologists rated FEAT's outputs as meeting or exceeding expert standards on most quality dimensions, confirming its practical utility and robustness in real forensic workflows. 
FEAT's design principles-modular transparency, continuous validation, and domain-specific adaptation-offer transferable insights applicable to other safety-critical AI applications, from clinical decision support to medicolegal reasoning, thereby addressing current limitations in human expertise and resource availability.

\begin{figure*}[htbp]
	\nolinenumbers
	\centering
	\captionsetup{font={tiny,bf,stretch=1.25},justification=raggedright}
	\includegraphics[width=\linewidth]{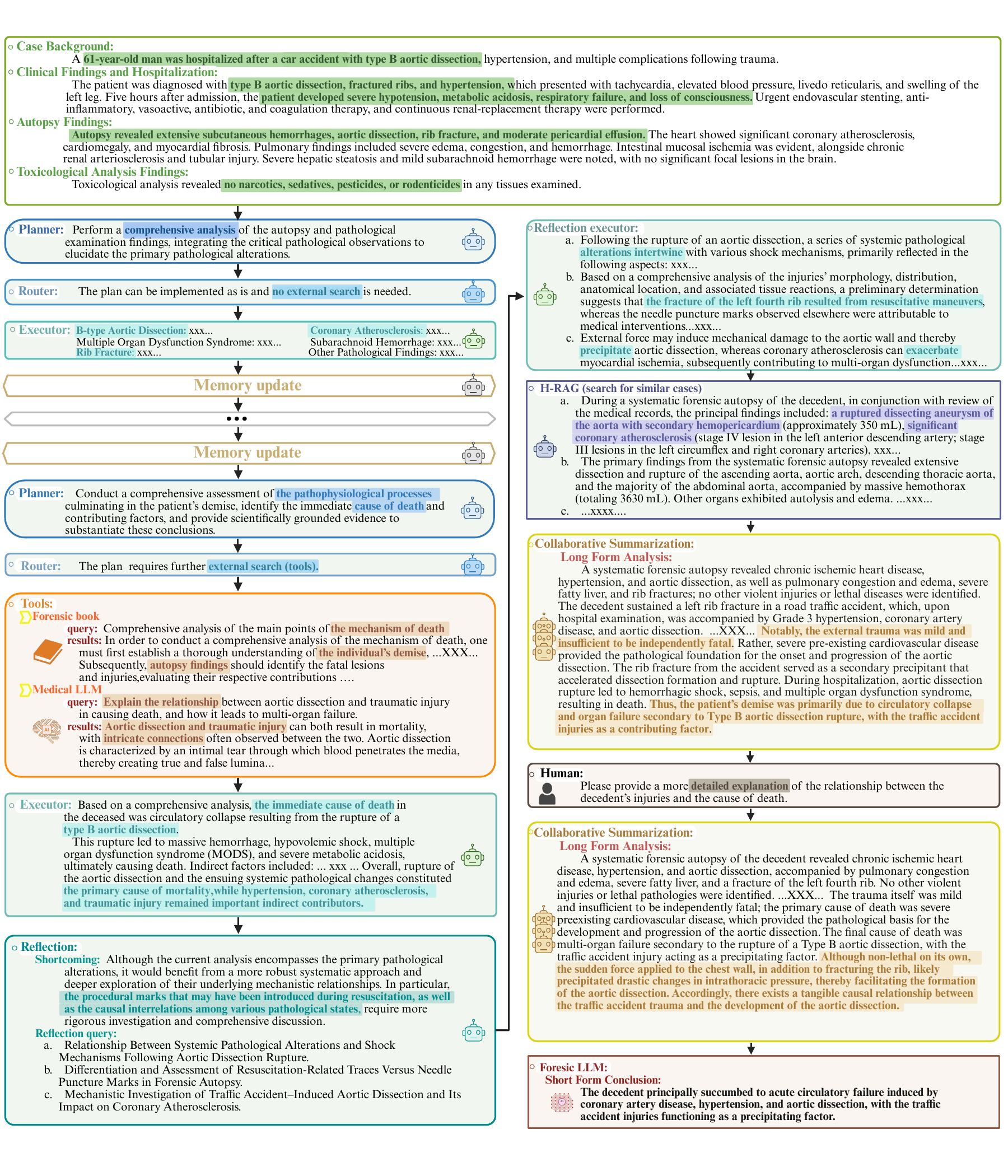}
	\caption{An example to illustrate the working process of FEAT for autonomous cause-of-death analysis. Specifically, given a real-world case (traffic injury with Type~B aortic dissection and chronic ischemic heart disease), 
	FEAT (i) ingests multi-source evidence (case background, clinical course, autopsy, toxicology); (ii) Planner decomposes the problem into subtasks and issues an execution plan; (iii) Local Solvers execute subtasks under a Router that decides between direct LLM inference and tool calls (e.g., forensic textbook vector store, medical LLM), yielding evidence-grounded intermediate findings; (iv) the coupled \textbf{Reflection \& Memory} module audits outputs for completeness/consistency, maintains a dynamic case file, and triggers replanning when gaps arise; and (v) the \textbf{Global Solver} performs collaborative summarization and invokes hierarchical RAG (\textbf{H-RAG}) to retrieve similar cases and authoritative references, then---optionally with human feedback---uses Forensic-LLM to produce court-ready conclusions (cause and manner of death) with traceable rationale. Arrows indicate the iterative Planner \textrightarrow Local Solvers \textrightarrow Memory \textrightarrow  Global Solver loop that continues until a coherent, evidence-grounded resolution is reached.}
	\label{fig:example}
\end{figure*}

\section*{Results} \label{sec:result}

\subsection*{FEAT for Autonomous Cause-of-Death Analyses}

FEAT employs a multi-agent framework (FIG.~\ref{fig:agent}) where specialized LLM-powered agents simulate forensic investigative roles through four coordinated components: (i) Planner, (ii) Local Solvers, (iii) Reflection $\&$ Memory, and (iv) Global Solver, operating in an iterative reasoning loop.

The \textbf{Planner} acts as the strategic orchestrator, analogous to a lead forensic pathologist. Upon receiving a case with multi-source information (e.g., scene summaries, autopsy reports, witness statements), it decomposes the analysis into subtasks (e.g., ``Assess poisoning indicators,'' ``Analyze traumatic injuries,'' ``Review medical history'') using CoT prompting to ensure comprehensive coverage of potential causes (trauma, toxicity, disease, etc.).

\textbf{Local Solvers} address these subtasks as specialized domain experts---for instance, an Autopsy Analyzer interpreting tissue pathology or a Toxicology Interpreter evaluating laboratory findings. Each solver implements the ReAct paradigm, augmenting LLM-driven reasoning with practical tool utilization, such as querying medical APIs or crime scene repositories. This generates evidence-based intermediate conclusions, exemplified by statements like, ``Autopsy reveals frothy fluid in airways and pulmonary edema, consistent with drowning,'' or ``Toxicology reports indicate a lethal concentration (3.0 µg/mL) of substance X.''

Intermediate outputs from Local Solvers populate a centralized \textbf{Memory} module, which serves as an integrated dynamic case file preserving contextual coherence beyond LLM prompt limits, thus ensuring comprehensive evidence retention despite task fragmentation.
A unique component of FEAT is the \textbf{Reflection} mechanism, closely coupled with Memory. Following accumulation of the Local Solvers' intermediate findings, the Reflection mechanism critically evaluates this evidence for internal consistency and completeness---addressing queries such as, ``Are all wounds accounted for?'', ``Do toxicology findings align logically with the presented scenario?'' and ``Is any evidence contradictory or overlooked?'' Should discrepancies or gaps be identified, the Reflection triggers revisions by the Planner. This Planner$\to$Local Solvers$\to$Reflection cycle iterates continuously until achieving a coherent, error-free resolution, mirroring the human forensic case review process. This self-correcting loop significantly enhances reliability and thoroughness, capturing and rectifying potential errors at early stages.

Once the evidence has been thoroughly analyzed and the Reflection yields no further questions, the \textbf{Global Solver} agent combines validated evidence with hierarchical RAG (H-RAG) and optionally human feedback, and then calls a forensically fine-tuned LLM to draft court-ready conclusions.
Typically, the format of generated conclusions is like: ``Cause of Death: Acute organophosphate poisoning leading to respiratory failure. Manner of Death: Suicide. Analysis: The victim's gastric contents and blood contained lethal levels of parathion; coupled with the absence of struggle or defensive wounds, this suggests self-ingestion of poison. No signs of foul play were detected. And ...''. Such conclusions are accompanied by an analysis explanation that references the key findings, thereby providing traceability from evidence to conclusion. In essence, the Global Solver plays the role of drafting the official autopsy report's conclusion section. This separation of planning and synthesis roles mirrors real-world forensic workflows, ensuring conclusions holistically weigh all evidence while maintaining auditability---a workflow comprehensively illustrated in FIG.~\ref{fig:example}.

\begin{figure*}[htbp]
	\nolinenumbers
	\centering
	\captionsetup{font={tiny,bf,stretch=1.25},justification=raggedright}
	\includegraphics[width=\linewidth]{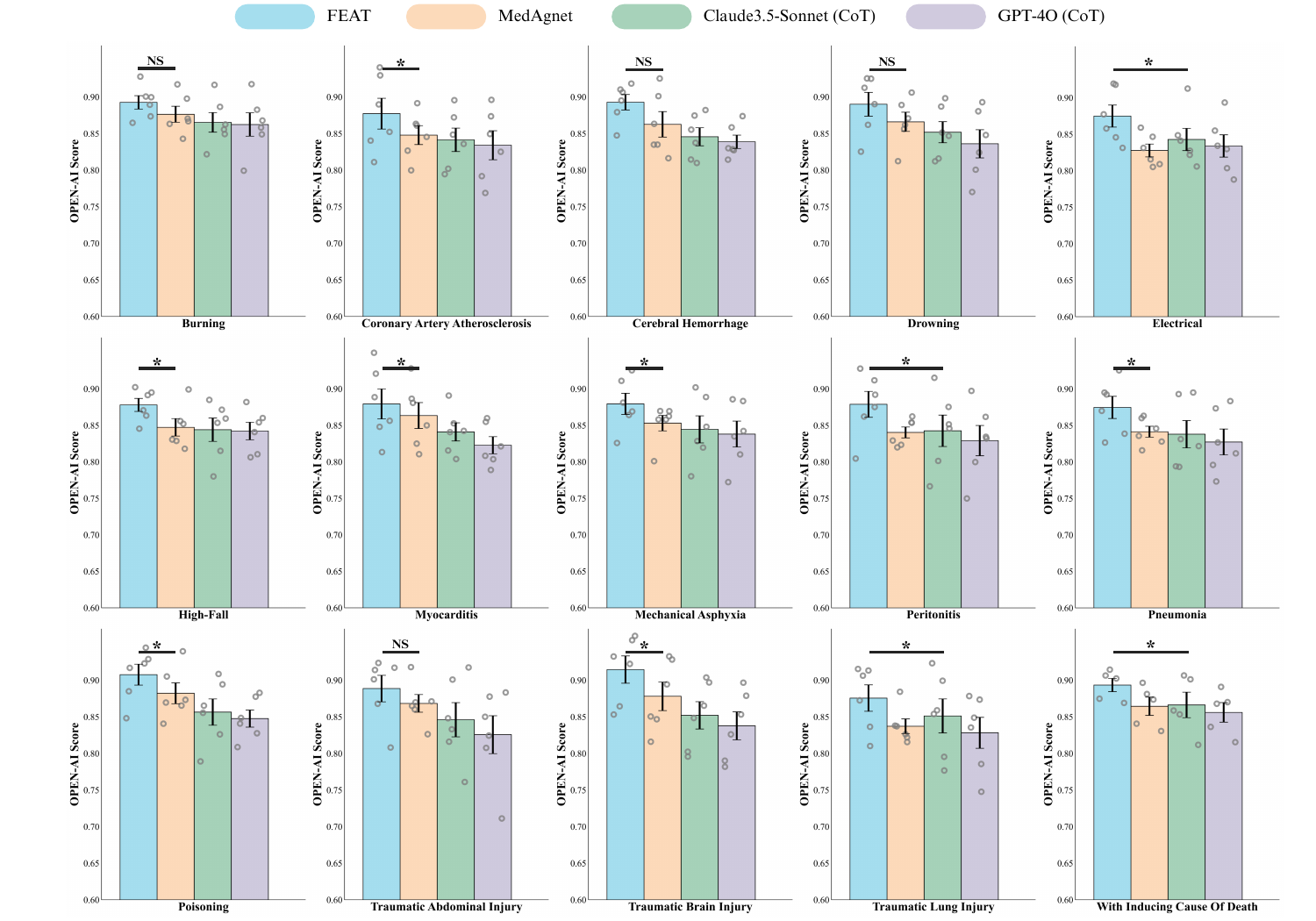}
	\caption{Long-Form Analysis (LFA) performance across 15 cause-of-death categories. Each panel reports the OpenAI-score (cosine similarity between embeddings of model outputs and expert references computed with text-embedding-3-large) for FEAT, MedAgent, Claude3.5-Sonnet (CoT), and GPT-4O (CoT). Bars denote mean\,$\pm$\,s.e.m. across cohorts; gray circles represent the average score of each cohort for that cause-of-death category. Horizontal lines and asterisks indicate one-sided Wilcoxon signed-rank tests of FEAT versus the best non-FEAT baseline.}
	\label{fig:analysis}
\end{figure*}

\subsection*{FEAT Significantly Improves Long-Form Analysis and Short-Form Conclusion Accuracy Across Diverse Cause-of-Death Categories}

Overall, FEAT was trained and evaluated on a Chinese-language medicolegal corpus comprising $7,748$ death investigation records collected from six leading forensic institutions across China. This rigorously curated dataset includes multimodal evidence streams, expertly annotated cause-of-death labels, and stratified training-test partitions for robust evaluation.
FEAT was rigorously evaluated against three state-of-the-art baselines --  \textbf{MedAgent}~\cite{medagent}, \textbf{Claude~3.5-Sonnet} (CoT)~\cite{claude}, and \textbf{GPT-4O} (CoT)~\cite{gpt4o}---across $15$ distinct cause-of-death categories. Performance was assessed using two key metrics: (i) \textbf{Long-Form Analysis (LFA)} quality, measuring comprehensive reasoning, and (ii) \textbf{Short-Form Conclusion (SFC)} accuracy, evaluating diagnostic precision, both quantified via the validated \textbf{OPENAI-score} metric (semantic similarity to expert-authored references).

As shown in FIG.~\ref{fig:analysis}, FEAT demonstrated consistent superiority in LFA tasks, achieving statistically significant improvements (Wilcoxon one-sided, p < 0.05) over the strongest baselines in $11$ of $15$ categories, prominently including complex multifactorial cases (i.e., Coronary Artery Atherosclerosis, Electrical Injury, High-Fall Trauma, Myocarditis, Mechanical Asphyxia, Peritonitis, Pneumonia, Poisoning, Traumatic Brain Injury, Traumatic Lung Injury, and deaths classified as With Inducing Cause of Death). These gains reflect FEAT's structured reasoning architecture and dynamic tool integration, which enhance multimodal evidence synthesis and pathophysiological interpretation. While four categories (Burning, Cerebral Hemorrhage, Drowning, Traumatic Abdominal Injury) did not reach significance, all exhibited positive performance trends, confirming FEAT's robust contextual reasoning even in diagnostically challenging scenarios.

\begin{figure*}[htbp]
	\nolinenumbers
	\centering
	\captionsetup{font={tiny,bf,stretch=1.25},justification=raggedright}
	\includegraphics[width=\linewidth]{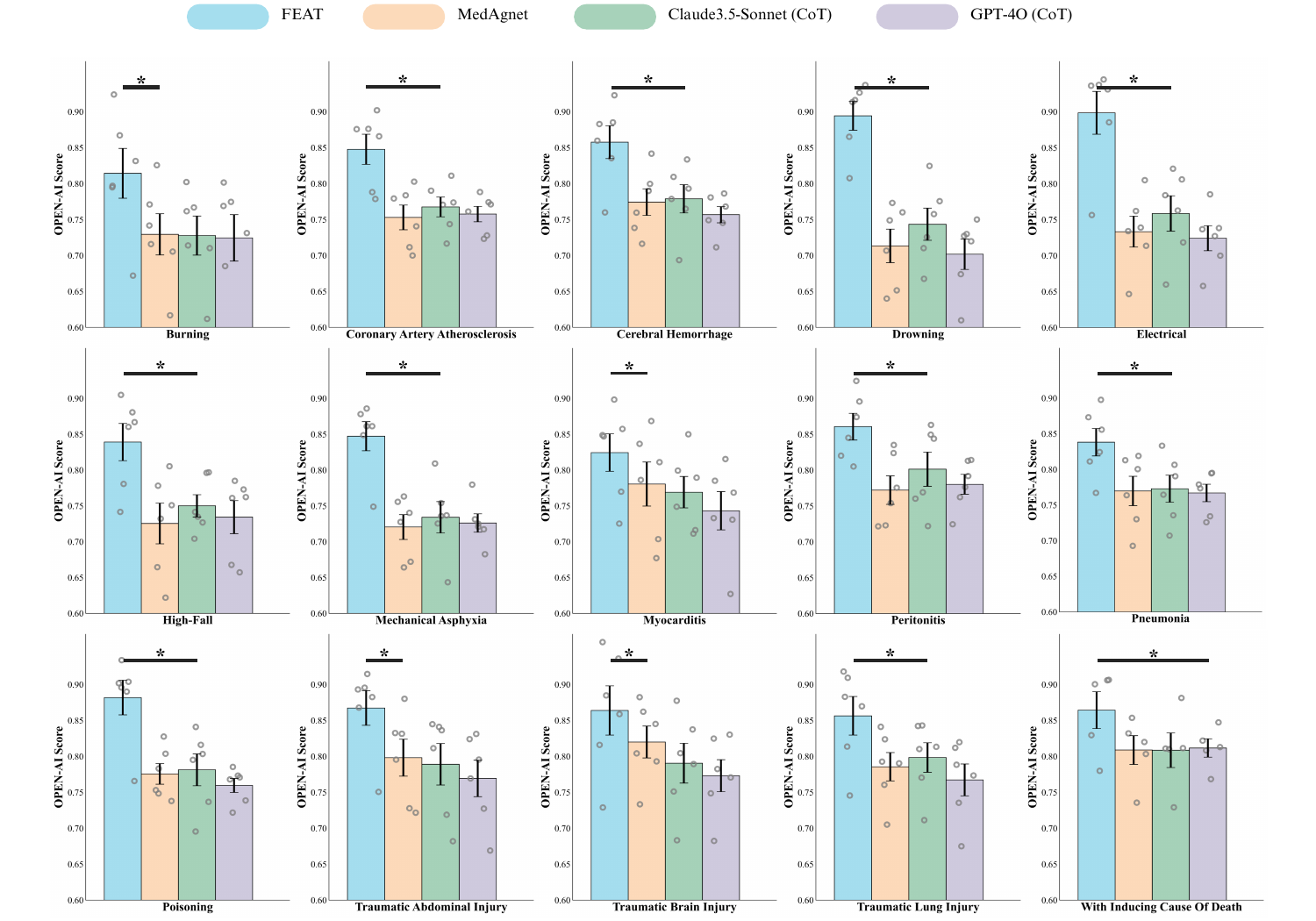}
	\caption{Short-Form Conclusion (SFC) performance across 15 cause-of-death categories. Each panel reports the OpenAI-score (cosine similarity between embeddings of model outputs and expert references computed with text-embedding-3-large) for FEAT, MedAgent, Claude3.5-Sonnet (CoT), and GPT-4O (CoT). Bars denote mean\,$\pm$\,s.e.m. across cohorts; gray circles represent the average score of each cohort for that cause-of-death category. Horizontal lines and asterisks indicate one-sided Wilcoxon signed-rank tests of FEAT versus the best non-FEAT baseline.}
	\label{fig:conclusion}
\end{figure*}

For the SFC task (FIG.~\ref{fig:conclusion}), FEAT achieved statistically significant superiority ($p < 0.05$) across all $15$ cause-of-death categories, generating concise, precise, and legally admissible cause-of-death determinations. These results demonstrate FEAT's dual capability: constructing comprehensive, reasonable, and coherent forensic narratives while distilling them into precise and legally actionable medicolegal conclusions. The performance advantage in SFC stems in part from domain-specific fine-tuning of the Forensic LLM, which aligns output structure and terminology with professional forensic standards.

These results show that FEAT significantly advances forensic pathology practice by simultaneously improving long-form analytical depth ($\Delta \, +3.2\%$ OPENAI-score in LFA) and short-form conclusion precision ($\Delta \, +10.7\%$ OPENAI-score in SFC) across all evaluated categories versus state-of-the-art benchmarks. These statistically validated improvements demonstrate enhanced coherence with expert reasoning patterns, directly strengthening the evidentiary robustness and legal admissibility of generated forensic opinions.

\subsection*{FEAT Demonstrates Robust Generalization Across Geographically Distinct Medicolegal Cohorts}

\begin{figure*}[htbp]
	\nolinenumbers
	\centering
	\captionsetup{font={tiny,bf,stretch=1.25},justification=raggedright}
	\includegraphics[width=\linewidth]{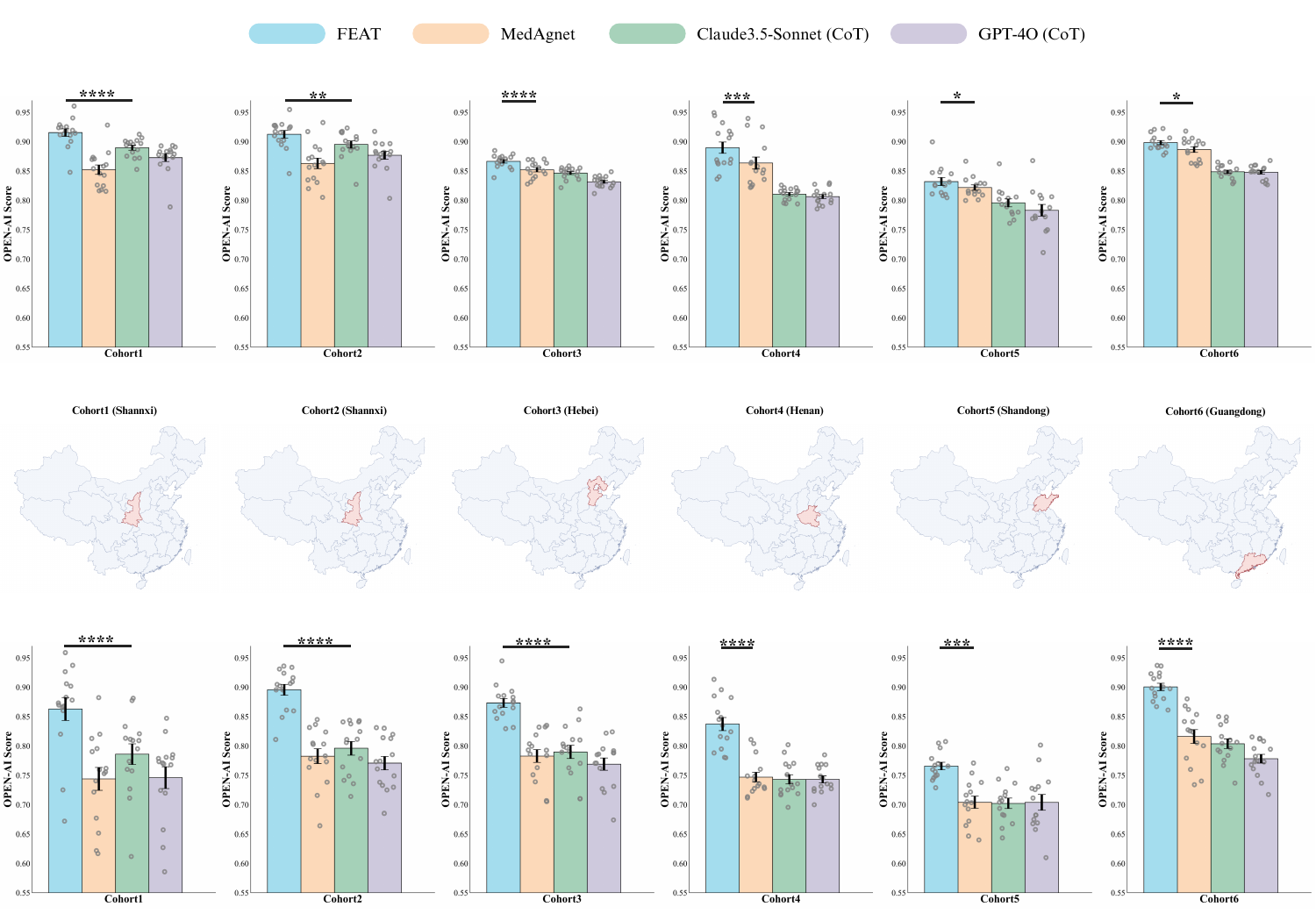}
	\caption{Geography-stratified performance of FEAT relative to leading AI baselines.
	For six independent provincial cohorts (maps, centre row), we quantified the conceptual concordance between each model's output and expert-annotated references using the OpenAI-score (cosine similarity of text-embedding-3-large model representations; higher is better).
Top row\enspace is the\enspace long-form analysis (LFA) quality. Bottom row\enspace is the short-form conclusion (SFC) accuracy.
Grey circles show per--cause-of-death scores; bars represent mean\,$\pm$\,s.e.m. Horizontal lines and asterisks indicate one-sided Wilcoxon signed-rank tests of FEAT versus the best non-FEAT baseline.}
	\label{fig:local}
\end{figure*}

Figure~\ref{fig:local} compares FEAT against three state-of-the-art models (MedAgent, Claude 3.5-Sonnet, GPT-4O) across \textbf{six distinct cohorts} drawn from geographically diverse Chinese provinces (2$\times$Shaanxi, Hebei, Henan, Shandong, Guangdong), each containing non-overlapping medicolegal cases stratified by 15 standardized cause-of-death categories. Per-category OPENAI-scores were plotted as vertical scatters superimposed on bar charts, and the accompanying error bars denote the cohort-level $mean \pm s.e.m.$; one-sided Wilcoxon signed-rank tests were applied within cohorts to assess statistical significance.
For LFA, FEAT achieved superior performance in all cohorts ($83.2\%$ to $91.5\%$ vs. the best-performing baseline $\Delta \, +1.2\%$-$3.0\%$, mean $+2.0\%$; all statistically significant with $p < 0.05$). The advantage expanded for SFC: \textsc{FEAT} recorded absolute scores of $76.5\%$-$90.0\%$ vs. $70.4\%$-$81.6\%$ for the best baseline, corresponding to a mean $\Delta\,+10.7\%$. All six comparisons remained significant after multiplicity correction ($p < 0.001$).
These consistent gains across provinces---despite variations in documentation practices and regional terminology---demonstrate FEAT's robustness. This resilience appears to stem from its hierarchical planning architecture and tool-augmented retrieval system, which together provide stable factual grounding, while the domain-adapted Forensic LLM and Reflection $\&$ Memory filtering module enhance alignment with local narrative styles and terminologies.

These results establish FEAT's consistent superiority over state-of-the-art multi-agent and CoT systems across previously unseen, geographically heterogeneous forensic datasets,  offering strong empirical support for its deployment at a national scale in medicolegal investigation workflows.

\begin{figure*}[htbp]
	\nolinenumbers
	\centering
	\captionsetup{font={tiny,bf,stretch=1.25},justification=raggedright}
	\includegraphics[width=\linewidth]{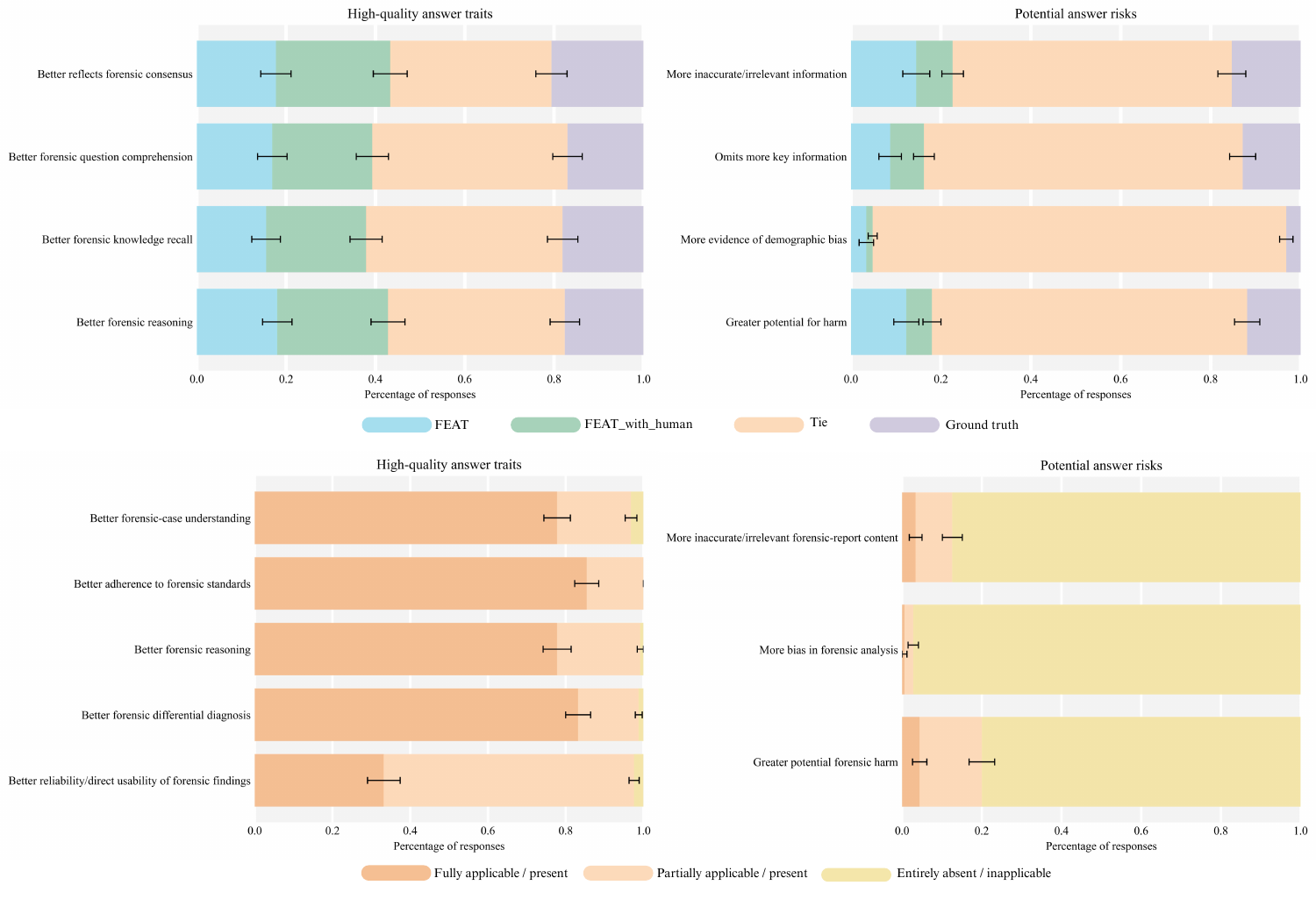}
	\caption{Expert appraisal of FEAT answer quality and safety.
	Top row\enspace --\enspace Blinded pair-wise preference tests.
	Four board-certified forensic pathologists compared stand-alone FEAT outputs and the human-in-the-loop variant with ground-truth reports for four desirable high-quality answer traits (left) and four potential answer-risk dimensions (right). Stacked bars show the proportion of comparisons in which each option was judged superior; beige segments denote ties.
Bottom row\enspace --\enspace Granular rating of final FEAT reports.
The same experts independently scored five positive attributes (left) and three risk factors (right) on a three-level rubric: fully applicable/present, partially applicable/present, and entirely absent/inapplicable. Error bars throughout represent two-sided 95 \% confidence intervals derived from $10^5$
non-parametric bootstrap resamples. }
	\label{fig:human}
\end{figure*}

\subsection*{Expert Evaluations Validate Forensic Reliability of FEAT}

A panel of four nationally recognized forensic pathologists (each with $>20$ years experience, from institutional affiliations spanning major Chinese forensic centers) thoroughly evaluated FEAT's forensic reliability from two aspects: (i) blinded comparisons of FEATS' outputs (both with and without human-in-the-loop) against expert-authored reports across eight forensic validity criteria, and (ii) granular rating its accuracy, applicability, and adherence to forensic standards. For more details, refer to the Method section.

The blinded pairwise comparisons (FIG.~\ref{fig:human}, upper panels) demonstrate three key findings. \emph{First}, standalone FEAT revealed comparable quality against expert reports, with frequent ``Tie'' rankings. Specifically, it scored approximately $2.5\%$  lower than expert reports on inaccurate/irrelevant information and key omissions, suggesting an intrinsic advantage of LLM-based agents in eliminating redundancy and capturing salient details. However, FEAT's risk scores for demographic bias and potential harm were marginally higher than those of the experts ($0.3\%$), highlighting the continuing need for human oversight to mitigate subtle biases and stylistic risks. \emph{Second}, for ``high-quality answer'' indicators (forensic consensus reflection, forensic question comprehension, forensic knowledge recall, and forensic reasoning), FEAT-with-human-in-the-loop outperformed both standalone FEAT ($+6.8\%$) and original reports ($+5.4\%$). 
It suggests that such an efficient review-feedback-revision paradigm markedly improves reasoning completeness and normative alignment. 
\emph{Third}, for ``potential risk'' indicators (inaccurate/irrelevant information, omission of key information, evidence of demographic bias, and harm potential), FEAT-with-human-in-the-loop exhibited lower adverse-event rates than both the standalone FEAT and the expert source texts, underscoring the value of human-AI collaboration in curbing factual errors, information gaps, and ethical hazards. 
These results collectively confirm that FEAT produces forensic-ready outputs meeting judicial standards, with human-AI collaboration further enhancing both accuracy and risk mitigation. 
The system's ability to maintain this performance across both quality and risk dimensions underscores its suitability for real-world medicolegal applications.

In fine-grained, single-model assessment (FIG.~\ref{fig:human}, lower panels), FEAT's unaided outputs were evaluated using an 8-criteria, three-tier rubric, including fully applicable/present, partially applicable/present, and entirely absent/inapplicable. 
The system demonstrated strong competency across the four high-quality answer dimensions: $77.8$-$85.4 \%$ of responses were rated fully applicable for forensic case understanding, standard adherence, reasoning quality, and differential diagnosis, with only a modest tail of partially applicable ratings. This indicates that FEAT reliably internalises professional guidelines, applies domain-specific logic, and articulates plausible alternative diagnoses at a level considered operationally sound by practising forensic pathologists. 
On the other hand, for the dimension of reliability or direct usability of forensic findings, enthusiasm was more tempered: one-third of responses ($33.2 \%$) were endorsed as fully usable, whereas nearly two-thirds ($64.4 \%$) were deemed only partially usable, suggesting that although the analytical core is robust, additional editorial refinement or contextual tailoring is still needed before full courtroom deployment.
Turning to potential answer risks, the modal rating for all three indicators---namely inaccurate or irrelevant content, bias in forensic analysis, and potential forensic harm---was entirely absent/inapplicable, comprising more than two-thirds of expert assessments, with high-risk occurrences limited to $< 5 \%$ (range: $0.6-4.4 \%$). It implies that FEAT seldom introduces substantive factual errors, demographic biases, or harmful recommendations capable of compromising legal proceedings. 
Overall, these results establish FEAT's capacity for high-fidelity forensic analyses with minimal safety liabilities, yet they also expose a practical gap in courtroom readiness that could be bridged through targeted post-editing or a light human-in-the-loop validation layer aimed at enhancing narrative reliability.

\begin{table}[htbp]
	\centering
	\nolinenumbers
	\caption{Ablation study of Forensic-LLM}
	\label{tab:tab1}
	
	\begin{tabular}{cc|ccccc}
		\toprule
		\multicolumn{2}{c|}{} & BLEU & ROUGE-1 & ROUGE-L & METEOR & OPENAI-Score \\ \midrule
		% Online LLM (API)
		\multirow{2}{*}{\parbox[c]{2cm}{\centering Online\\LLM (API)}}
		& GPT-4O\cite{gpt4o}          & $0.2464\pm0.2068$ & $0.4457\pm0.1814$ & $0.4190\pm0.1832$ & $0.3632\pm0.2166$ & $0.7929\pm0.0944$ \\
		& DeepSeek\cite{deepseek-v3}         & $0.3018\pm0.2452$ & $0.4729\pm0.2103$ & $0.4465\pm0.2148$ & $0.4005\pm0.2487$ & $0.8097\pm0.0968$ \\ \cmidrule{1-2}
		% Local fine-tuned LLMs
		\multirow{4}{*}{\parbox[c]{2cm}{\centering Local\\LLM\\fine-tuning}}
		& Qwen2-FT\cite{qwen} (7 B)  & $0.3440\pm0.2718$ & $0.5485\pm0.2178$ & $0.5193\pm0.2233$ & $0.4819\pm0.2517$ & $0.8278\pm0.0944$ \\
		& Llama-3-FT\cite{llama} (8 B) & $0.3744\pm0.2805$ & $0.5737\pm0.2243$ & $0.5402\pm0.2316$ & $0.5030\pm0.2559$ & $0.8341\pm0.1037$ \\
		& GLM-4-FT\cite{glm} (9 B)   & $0.3845\pm0.2760$ & $0.5802\pm0.2306$ & $0.5474\pm0.2403$ & $0.5140\pm0.2643$ & $0.8385\pm0.1044$ \\
		& DeepSeek-FT\cite{deepseek-v3} (8 B)& $\bm{0.3962\pm0.2875}$ & $\bm{0.5863\pm0.2255}$ & $\bm{0.5583\pm0.2336}$ & $\bm{0.5297\pm0.2564}$ & $\bm{0.8394\pm0.1048}$ \\ \bottomrule
	\end{tabular}
\end{table}

\subsection*{Modular Ablation Studies Validate FEAT's Optimized Forensic Analysis Architecture}

We conducted three ablation studies to validate FEAT's core design components: (i) external tool integration, (ii) planner reasoning depth, and (iii) forensic LLM selection (Table~\ref{tab:tab1}, \textit{Supplementary} Tables~3--6).

\textbf{Tool Integration}: We assessed the functional contribution of external tool utilization by fixing the planner agent outputs and disabling the tool agent in the local-solver (\textit{Supplementary} Tables~3 and 4). This ablation redirected all planner-generated steps to be executed directly by the executor without invoking any auxiliary analytical tools. The results unequivocally demonstrated that tool integration consistently augmented performance across all cause-of-death categories. For LFA, the incorporation of external tools yielded an average improvement of $0.020$ in the OPENAI-score (ranging from $0.002$ to $0.047$), with the most significant gains observed in Traumatic Brain Injury cases. Similarly, for SFC, the average performance uplift was $0.023$ (ranging from $0.000$ to $0.069$). 
This confirms tool-augmented reasoning's critical role in complex case analysis.

\textbf{Reasoning Depth}: To evaluate the influence of reasoning granularity, we experimented with planner depths of $2$, $4$, and $5$ steps (\textit{Supplementary} Tables~5 and 6). Results revealed a pronounced improvement in performance when increasing reasoning depth from $2$ to $4$. Specifically, depth-4 led to average OPENAI-score gains of $0.039$ ($0.023$--$0.061$) for LFA and $0.046$ ($0.003$--$0.097$) for SFC across all categories. However, further deepening the reasoning to depth-$5$ introduced diminishing returns and, in the majority of categories ($12$ out of $15$), resulted in performance degradation. This regression is likely attributable to over-decomposition, where excessive CoT expansion introduces cognitive noise, redundancy, and semantic drift. 
This establishes depth-4 as the precision-efficiency optimum.

\textbf{LLM Selection}: We further investigated the role of LLM and domain adaptation in SFC (Table~\ref{tab:tab1}). Six LLMs were evaluated: two commercial online models (GPT-4O and DeepSeek) and four domain-adapted local models (Qwen2, LLaMA3, GLM4, and DeepSeek). Each model received identical long-form analyses as input and generated ten short-form conclusions per case, from which average scores across five forensic evaluation metrics were computed. The results demonstrate a clear advantage for domain-specific adaptation: all four locally fine-tuned models surpassed the commercial baselines, affirming the efficacy of targeted knowledge infusion. Among them, the locally adapted DeepSeek model consistently achieved the highest scores across all metrics, establishing it as the most performant forensic-LLM for integration into FEAT's architecture.

These ablation results collectively validate FEAT's architectural triad. The integration of external analytical tools, a planner configured with calibrated reasoning depth, and the adoption of domain-specialized LLMs together contribute to significant and measurable improvements in the quality, precision, and reliability of forensic cause-of-death determinations.

\section*{Discussion} \label{sec:discussion}

This work introduces FEAT, a novel multi-agent framework that bridges AI and forensic medicine through domain-adapted LLMs for automated cause-of-death analysis. 
Empirically, the system elevates the accuracy and consistency of death certification beyond traditional, largely manual practices: its LLM-driven agents can parse heterogeneous forensic data, reason over complex evidence, and reliably infer the underlying cause of death, thereby strengthening public-health surveillance and forensic pathology.
For practitioners, FEAT serves as an expert-level decision-support system that processes case materials and generates autopsy reports, enhancing investigation efficiency, standardization, and nationwide quality of forensic examinations---an area where AI adoption has historically lagged behind other medical domains~\cite{gap}. 
From the standpoint of AI research, FEAT pushes the frontier from static question-answering toward full, high-stakes problem-solving: its multi-agent, tool-enhanced architecture and built-in reflection mechanisms illustrate how LLMs can be orchestrated into transparent, domain-aware systems capable of complex reasoning. 
These contributions advance both forensic medicine and AI by suggesting a safe, effective pathway for integrating LLMs into expert decision processes.

%Overcoming Challenges in Forensic Pathology
FEAT's multi-agent architecture directly targets four long-standing impediments to modern forensic practice.
\textit{First}, excessive caseloads---pathologists globally exceed recommended autopsy volumes, with particular severity in China, where urban centers face concentrated mortality while rural regions lack adequate resources~\cite{china_3,china_4,workforce_size}.
FEAT automates comprehensive report generation and critical finding identification, alleviating case processing bottlenecks while enabling experts to focus on complex or contentious investigations.
\textit{Second}, regional quality disparities have historically produced inconsistent determinations between well-equipped metropolitan universities and less-developed jurisdictions~\cite{diversity,wrong_1,wrong_2}. Evaluated across a broad, nationwide Chinese case corpus, FEAT supplies a standardised analytical workflow that generalises across local practice patterns, enabling less-experienced examiners to reach diagnostic accuracies that match up with senior specialists.
\textit{Third}, cause-of-death reasoning demands the integration of heterogeneous evidence (e.g., autopsy records, scene reports, witness statements, and toxicology data), typically presented in multiple formats and at variable granularity~\cite{information_1,information_2,information_3,information_4}.
Capitalizing on domain-adapted LLMs, FEAT parses multimodal inputs, deploying specialized agents for evidence abstraction, hypothesis generation, and differential reconciliation to construct coherent explanatory narratives---surpassing the limitations of rule-based and statistical systems.
\textit{Fourth}, FEAT prioritizes transparency by revealing its full chain of reasoning, enabling examiners to scrutinize every step and ensuring cause-of-death determinations remain objective and fully auditable.

%Comparison to Prior AI Approaches and Unique Architecture
Prior forensic AI systems relied primarily on narrow classifiers~\cite{classification} and wide-and-deep networks~\cite{fpath,songci} that generated silent predictions from structured inputs, lacking explanatory reasoning capabilities.
In contrast, FEAT introduces an end-to-end, narrative-generating pipeline whose reasoning can be read, audited, and defended.
While traditional medical NLP systems extracted fragmented text elements~\cite{clinical_bert,clinical_gpt}, none achieved end-to-end analysis from raw case materials to legally admissible conclusions---a gap FEAT bridges by generating expert-level forensic reports.
Directly deploying a single, general-purpose LLM (e.g., GPT-4O or DeepSeek) would lead to unreliable forensic analysis due to factual hallucinations, poor multi-step reasoning, and lack of domain grounding~\cite{hallucination}. FEAT addresses these limitations by coordinating specialized, forensically-constrained LLM multi-agents through structured ``case conference'' protocols.
Dedicated agents perform distinct forensic functions (e.g., autopsy abstraction, scene verification, and hypothesis testing), replicating human peer-review through computational debate and correction---an approach inspired by emerging multi-agent LLM research (e.g., the ``Stanford Town''~\cite{AI_town} and ``AI Hospital''~\cite{AI_hospital}) but applied here for the first time to forensic medicine.
Crucially, every agent can call external tools: querying forensic domain databases, understanding post-mortem knowledge, or retrieving the latest guidelines, thereby grounding its reasoning in up-to-date facts and overcoming the knowledge-cut-off limitations of static models. 
Built-in reflection mechanisms and human-in-the-loop verification jointly detect inconsistencies and refine conclusions---leveraging iterative CoT and self-correction to enhance reasoning accuracy.
Together, multi-agent debate, tool augmentation, and iterative self-correction yield a transparent audit trail: each inference step is logged, cited to specific evidence, and reproducible, satisfying the stringent explainability demands of medicolegal practice.

%Practical Benefits in Chinese Forensic Workflows
In practical deployment across Chinese medicolegal workflows, FEAT exhibits a spectrum of operational benefits that translate directly into higher-quality casework and institutional efficiency. 
FEAT generates long-form forensic analyses comparable to expert pathologists' reports---demonstrating thorough, well-structured explanations that account for critical findings. These high-quality drafts reduce documentation time while serving as real-time quality controls, automatically flagging omissions for human review to enhance accuracy in routine cases. 
FEAT delivers precise short-form conclusions for death certificates and legal filings, reliably identifying immediate causes, contributory factors, and manner of death in close agreement with human experts. It captures critical distinctions (e.g., accident vs. homicide, disease vs. trauma, underlying/inducing vs. immediate/direct causes of death), enhancing both individual case accuracy and aggregate mortality statistics.
Critically, FEAT generalizes across China's diverse regions, adapting to dialectal variations, local practices, and crime patterns with consistent accuracy. Its multi-agent architecture activates case-specific reasoning (violent crime vs. natural disease vs. industrial accidents), ensuring equitable performance across urban and rural jurisdictions and mitigating historical resource disparities.
FEAT’s reasoning aligns with expert forensic judgment through pathologist-guided fine-tuning, ensuring cited factors (injury severity, disease markers, scene context) match human diagnostic logic. Expert evaluations confirm its transparent, forensically plausible outputs serve as a didactic aid for trainees.
FEAT integrates seamlessly with existing digital infrastructure, processing inputs from standard information systems or transcribed notes and delivering practitioner-friendly outputs. This AI-assistant pre-processes evidence, highlights critical findings, and generates draft reports in $<10$ minutes, which is particularly crucial during workload surges. By accelerating case throughput while safeguarding against oversights, FEAT has the potential to enhance both productivity and quality across China's medicolegal system.

% limitation

FEAT demonstrates significant advantages, but several limitations highlight the need for further research before global adoption can be achieved. \emph{First}, the system's current design is specific to Chinese contexts, with language models, knowledge bases, and forensic standards optimized for domestic use. This localization may limit its performance when applied to foreign languages or different legal frameworks. \emph{Second}, human oversight remains essential due to the high-stakes nature of cause-of-death determinations; while FEAT achieves strong accuracy, occasional errors necessitate expert validation to maintain reliability and public trust. \emph{Third}, the system has not yet met China's stringent legal requirements for courtroom testimony or death certification, requiring extensive field trials and potential regulatory reforms. \emph{Fourth}, interpretability challenges persist despite FEAT's reasoning logs, as the complexity of multi-agent interactions may obscure subtle flaws in analysis. Improved methods for linking inferences to supporting evidence will be crucial for legal scrutiny. \emph{Finally}, potential biases from historical case imbalances must be addressed through continuous monitoring and human review of sensitive outputs. 
In a nutshell, these limitations chart a clear future agenda: (i) extending FEAT through cross-lingual and cross-jurisdictional fine-tuning, (ii) embedding stronger guardrails and validation protocols, (iii) advancing interpretable-AI techniques tailored to forensic reasoning, and (iv) instituting rigorous bias-detection pipelines.
Together, these steps will enable FEAT's evolution from a China-specific system to a universally reliable forensic platform.

To sum up, FEAT bridges forensic science and AI by demonstrating how carefully designed multi-agent systems---integrating domain adaptation, tool augmentation, and memory mechanisms---can achieve near-expert performance in complex, specialized tasks. For forensic pathology, it addresses critical challenges of expertise scalability and procedural consistency, while offering the AI community a framework for deploying language models in high-stakes domains. Although limitations require cautious integration, FEAT's results establish its potential as a transformative adjunct in medicolegal investigations. Through continued refinement and human-AI collaboration, such systems may redefine forensic practice by combining human expertise with AI's analytical consistency.

\section*{Methods} \label{sec:method}

\subsection*{The Multi-Agent Framework of FEAT}

\textsc{FEAT} is a closed-loop, multi-agent system that converts heterogeneous forensic evidence into a transparent and legally defensible cause-of-death determination.  
A fine-grained task-disassembly \textbf{Planner} decomposes each case into a hierarchical reasoning tree, automatically selects diagnostically pivotal nodes, and re-orders them into an optimised execution plan, which is then routed to \textbf{Local Solvers}.  
For every task, a Router agent decides whether direct LLM inference suffices or whether external augmentation is required; when augmentation is necessary, a ReAct-based Tool agent can query web search, consult a curated vector store of forensic textbooks, retrieve PubMed abstracts, or invoke a domain-specific medical LLM, after which an Executor synthesises evidence-grounded answers.
All intermediate outputs flow into the \textbf{Reflection \& Memory} module, where a structured filter prompt suppresses redundancy and hallucinations, verifies consistency against fixed background facts, and condenses validated findings into length-bounded summaries that populate a low-entropy, long-horizon memory accessible to every agent.
Finally, a \textbf{Global Solver} applies a two-granularity hierarchical retrieval-augmented generation mechanism-operating on both sentence- and paragraph-level embeddings---to retrieve stylistic exemplars from a corpus of $6,739$ expert analyses; a Collective-Summarisation agent fuses these exemplars with the validated Local-Solver outputs and any optional human critiques, and passes the refined context to a LoRA-adapted forensic LLM, which produces a detailed analytic report and a concise, court-ready conclusion (Fig.~\ref{fig:agent} and Fig.~\ref{fig:example}).  
By combining hierarchical planning, tool-enhanced reasoning, memory consolidation, controlled RAG, and a locally fine-tuned forensic LLM within a single architecture, \textsc{FEAT} provides transparent, reproducible, and scalable medico-legal decision support.

\subsubsection*{Planner Agent}

Building on the ToT~\cite{tot} and SELF-DISCOVER~\cite{self-discover} frameworks, our planner forms a dynamic module that systematically explores, prunes, and optimises reasoning paths for forensic cause-of-death analysis. 
For each task specification $T$, the planner constructs a hierarchical reasoning tree $\mathcal{T}=(\mathcal{V},\mathcal{E})$ whose depth is adaptively chosen by a LLM but capped at four levels; nodes $\mathcal{V}$ represent individual reasoning steps, and edges $\mathcal{E}$ encode their logical dependencies.
The planner architecture encompasses three distinct stages: Tree Construction (Plan Generation), Node Selection (Key Step Selection), and Node Adaptation (Structure Optimization). 

In the first stage, the planner decomposes a complex forensic task into smaller, manageable reasoning steps, forming a reasoning tree. Given an initial forensic problem description $x$, a large language model $\mathcal{M}$ iteratively generates multiple reasoning paths, represented formally as:
\begin{equation}
 \mathcal{T} = \mathcal{M}(p_{tree},x) \rightarrow \mathcal{T}(\mathcal{V},\mathcal{E}),
\end{equation}
where $p_{tree}$ is a structured prompt instructing the LLM to decompose the task into hierarchical subtasks(\textcolor{black}{see Supplementary Fig.~1}). Each node 
$v \in \mathcal{V}$ is a coherent reasoning step described by a textual instruction (e.g., ``evaluate autopsy results'', ``assess toxicology report'', ``integrate multimodal data'').

In the node selection stage,  the planner autonomously identifies key branching nodes within the reasoning tree without pre-defined manual thresholds, inspired by the SELF-DISCOVER framework.
We package the entire reasoning tree \(\mathcal{T}\) as
$\bigl\{\,\bigl(v,\;\mathrm{subtree}(v)\bigr)\bigr\}_{v\in\mathcal{V}}$, and invoke a single meta-prompt \(p_{\mathrm{select}}\) that directs the model to consider each step's relevance, diagnostic impact, and depth of forensic insight.  Concretely, the LLM call is:
\begin{equation}
\mathcal{V}^* 
\;=\; 
\mathcal{M}\!\Bigl(
p_{\mathrm{select}},\;
\{(v,\mathrm{subtree}(v))\}_{v\in\mathcal{V}}
\Bigr),
\end{equation}
where \(p_{\text{select}}\) is the meta-prompt instructing the LLM to assess node relevance, significance, and informativeness about the forensic task and \(\mathcal{V}^*\subseteq\mathcal{V}\) is the returned subset of nodes deemed essential for deeper analysis(\textcolor{black}{see  Supplementary Fig. 2}).  This design leverages the model's intrinsic comparative reasoning and obviates any explicit numeric aggregation or external decision rules.

In the final adaptation phase, each selected node and its corresponding subtrees undergo further refinement and optimization.
Given the autonomously selected critical nodes \(\mathcal{V}^{*}\subseteq\mathcal{V}\), the plan adaptation is performed individually for each node \(v^{*}\in\mathcal{V}^{*}\).
\begin{equation}
	v^{*}_{\text{adapted}} = \mathcal{M}(p_{\text{adapt}}, v^{*}, \text{subtree}(v^{*})),
\end{equation}
where \(p_{\text{adapt}}\) is the meta-prompt explicitly guiding the model to refine and optimize the reasoning structure of each node and subtree for forensic-specific analytical requirements(\textcolor{black}{see  Supplementary Fig. 2}).
\(\text{subtree}(v^{*})\) refers to the detailed reasoning steps following from node \(v^{*}\), providing comprehensive context essential for optimization.
The complete adapted reasoning tree/chain \(\mathcal{T}_{\text{adapted}}\) is thus obtained by applying the adaptation across all selected critical nodes:
\begin{equation}
	\mathcal{T}_{\text{adapted}} = \{ v^{*}_{\text{adapted}} \mid v^{*}\in \mathcal{V}^{*} \}.
\end{equation}
\textcolor{black}{is then emitted as a sequence of actionable instructions for the Local Solver agents.  This three-stage design allows the planner to explore divergent lines of enquiry while converging on a compact set of high-value reasoning steps tailored to the evidentiary nuances of each case, thereby ensuring clarity, efficiency, and robustness in downstream analysis.}

\subsubsection*{Local Solver Agent}

\textcolor{black}{The Local Solver translates the planner’s high-level reasoning steps into concrete forensic actions through three tightly coupled sub-agents--\textbf{Router}, \textbf{Tool}, and \textbf{Executor}--that handle dynamic task routing, external tool invocation, and final evidence-grounded reasoning.}

The Router Agent acts as the central decision-making unit that determines the complexity and tool dependency of each sub-task generated by the Planner. Given a planner-generated task:
\begin{equation}
	\tau \;=\;\bigl\{\texttt{task\_instruction},\; \texttt{forensic\_background}\bigr\},
\end{equation}
where \texttt{task\_instruction} encodes the sub‑goal and \texttt{forensic\_background} contains case-specific data (e.g., pathology, toxicology), the Router decides whether external tools are needed(\textcolor{black}{see Supplementary Fig. 3}).  
We implement the decision rule as an LLM-based classifier:
\begin{equation}
\mathcal{R}(\tau)=
\begin{cases}
	0 & \text{if direct inference suffices},\\[4pt]
	1 & \text{if external tool augmentation is required}.
\end{cases}
\end{equation}
\begin{itemize}
	\item $\mathcal{R}(\tau)=0$: \(\tau\) is forwarded directly to the \textbf{Executor Agent}.
	\item $\mathcal{R}(\tau)=1$: \(\tau\) is delegated to the \textbf{Tool Agent}.
\end{itemize}

The Tool Agent follows the ReAct paradigm~\cite{react}, iterating over \emph{Thought}~$\to$~\emph{Tool Selection}~$\to$~\emph{Tool Execution}~$\to$~\emph{Observation}.  
Let \(h_{0}\) be the initial dialogue\text{-}plus\text{-}memory state.  
We define:
\begin{equation}
T:\mathcal{H}\!\to\!\mathcal{T},\;
S:\mathcal{T}\!\to\!\mathcal{S},\;
U:\mathcal{S}\!\to\!\mathcal{E},\;
O:\mathcal{E}\!\to\!\mathcal{O},
\end{equation}
where \(\mathcal{H}\) is the history space, \(\mathcal{T}\) latent thoughts, \(\mathcal{S}\) tool specs, \(\mathcal{E}\) raw tool outputs, and \(\mathcal{O}\) processed observations.  
For \(k=1,\dots,K\):
\begin{align}
	\tau_{k} &= T(h_{k-1}) &&\text{(thought)}\\
	s_{k}    &= S(\tau_{k}) &&\text{(tool select)}\\
	r_{k}    &= U(s_{k})    &&\text{(tool exec.)}\\
	o_{k}    &= O(r_{k})    &&\text{(observe)}\\
	h_{k}    &= \text{concat}(h_{k-1},o_{k}) &&\text{(update)}
\end{align}
Compactly,
\begin{equation}
h_{k}=\text{concat}\bigl(h_{k-1},(O\!\circ\!U\!\circ\!S\!\circ\!T)(h_{k-1})\bigr).
\end{equation}
After \(K\) iterations, the enriched state \(h_{K}\) is sent to the Executor.

In our system, we integrate four external modules to support information retrieval and domain-specific reasoning:
\begin{enumerate}
	\item \textbf{Website Search Tool} : 
	We employ the Tavily Search API~\footnote{\url{https://tavily.com/}} to perform general‐purpose web queries. Tavily abstracts crawling, scraping, filtering, and ranking, returning up‑to‑date, citation‑ready results optimized for integration with LLMs and autonomous agents.
	
	\item \textbf{Domain Knowledge DataBase} : 	
	We constructed a curated vector store by segmenting authoritative forensic medicine textbooks into coherent passages using a BERT‑based semantic segmenter~\cite{seg-bert}. Each passage was embedded via OpenAI's text embedding model~\cite{embedding} and indexed in a Chroma vector database~\footnote{\url{https://www.trychroma.com/}}. At inference, user queries are embedded and matched by cosine similarity to retrieve the top‑$k$ ($k=2$) most relevant passages, providing a lightweight retrieval\text{-}augmented generation (RAG) capability.

	\item \textbf{PubMed Dataset Retriever} : 	
	We utilize LangChain's PubMed integration~\footnote{\url{https://python.langchain.com/docs/integrations/retrievers/pubmed/}} to search MEDLINE and PubMed Central. Given a medical or forensic query, the retriever returns a ranked list of peer‑reviewed articles ($num=3$ in our study), often with direct links to full texts, thus supporting evidence‑based model outputs.

	\item \textbf{Medical LLM}: We integrate Baichuan-M1~\cite{baichuan}, a family of LLMs trained from scratch on 20 trillion tokens with a dedicated medical focus. Unlike approaches that simply continue pretraining on general models, Baichuan-M1's architecture and a data pipeline are explicitly optimized for clinical and medical tasks, yielding strong performance across both broad and specialized benchmarks. In our system, we encapsulate the model's inference endpoint with a single, well-defined function interface. This abstraction presents Baichuan-M1 as a self-contained tool that downstream components can invoke via a uniform prompt-response call.
	
\end{enumerate}

The Executor Agent, denoted by $E$, is implemented as a large‑capacity language model that is queried with a structured prompt $p_{executor}$(\textcolor{black}{see  Supplementary Fig. 4}).  This prompt contains three logically distinct fields: the \text{task instruction} and \text{forensic background} $\tau_i$, the optional tool\text{-}augmented evidence $O_i$, and the cumulative memory $M_{i-1}$.
Because \(O_i\) and \(M_{i-1}\) can exceed the context window of \(E\), a lightweight summarisation model \(S_{\text{sum}}\) compresses both sources while preserving task\text{-}relevant semantics.  
Consequently, the Executor receives the distilled prompt \(\{\tau_i,\, S_{\text{sum}}(O_i),\, S_{\text{sum}}(M_{i-1})\}\) and produces the local result:
\begin{equation}
R_{\text{local}_i} \;=\;
E\!\bigl(p_{\text{executor}},\, \tau_i,\, S_{\text{sum}}(O_i),\, S_{\text{sum}}(M_{i-1})\bigr).
\end{equation}

\textcolor{black}{which is appended to the Reflection-and-Memory buffer, closing the perception--action loop for subsequent planning cycles.
This tri-agent design enables fine-grained control over task complexity, seamless integration of heterogeneous knowledge sources, and context-aware synthesis of forensic evidence, thereby ensuring clarity, robustness, and reproducibility in downstream medico-legal reasoning.}

\subsubsection*{Global Solver Agent}

\textcolor{black}{The Global Solver orchestrates document-level synthesis through a three-stage, retrieval-augmented pipeline that combines \emph{hierarchical retrieval-augmented generation} (H-RAG), \emph{gap-oriented collective summarisation}, and a domain-adapted \emph{Forensic-LLM} fine-tuned with Low-Rank Adaptation (LoRA)~\cite{lora}.}

A corpus of $N = 6{,}739$ expert-written cause-of-death analyses,
\(
\mathcal{D}=\{y_i\}_{i=1}^{N},
\)
contains no evidentiary details and thus serves exclusively as a stylistic reference.  
To mimic the multi-level reasoning of human experts, H-RAG operates on two semantic granularities: each Local-Solver sentence $s_j$ and each paragraph-level memory summary $p_k$ is embedded with \texttt{text-embedding-3-large}, yielding vectors,
\(
\mathbf{s}_j=\operatorname{emb}(s_j)
\)
and
\(
\mathbf{p}_k=\operatorname{emb}(p_k).
\)

Each embedding $\mathbf{z} \in { \mathbf{s}_j } \cup { \mathbf{p}_k }$ is matched against the corpus $\mathcal{D}$ via cosine similarity to identify its two nearest neighbors:
\begin{equation}
	T(\mathbf{z})\;=\;\operatorname{Top}_{2}(\mathcal{D},\mathbf{z})
	\;=\;\underset{\substack{\mathbf{y}_i\in\mathcal{D}\\|T|=2}}%
	{\arg\max}\;
	\cos\!\bigl(\mathbf{z},\operatorname{emb}(y_i)\bigr).
\end{equation}

The set of retrieved candidates from all query vectors is aggregated into a multiset $\mathcal{L} = \bigcup_{\mathbf{z}} T(\mathbf{z})$. To ensure global consensus, we define a frequency score:
\(
f(y_i)=\bigl|\{\mathbf{z}\mid y_i\in T(\mathbf{z})\}\bigr|,
\)
and retain the top four most frequently retrieved corpus items:
\begin{equation}
	\mathcal{Y}^{\star}=\operatorname{Top}_{4}\bigl(\mathcal{D},f\bigr),
\end{equation}

This hierarchical retrieval procedure---``top-2-per-query, top-4-by-consensus''---balances local relevance with corpus-wide representativeness, yielding a succinct, exemplar-based reference set $\mathcal{Y}^{\star}$ used as in-context examples for few-shot generation.

Let $R = \{r_1, \ldots, r_m\}$ denote the Local Solver's initial conclusions. A Collective Summarization Agent then identifies content omissions or inconsistencies by producing a query set 
\begin{equation}
	Q = \{ q_1, \ldots, q_K \} = \mathcal{M}( p_{\text{gap}}, R)
\end{equation}
using a structured gap-analysis prompt $p_{\text{gap}}$ provided to the LLM $\mathcal{M}$(\textcolor{black}{see  Supplementary Fig. 5}). Each query $q_k$ is recursively re-routed to the Local Solver, producing an expanded evidence set $R' = \{r'_1, \ldots, r'_n\}$. The final forensic analysis is generated as:
\begin{equation}
	A=\mathcal{M}\!\bigl(
	p_{\text{summary}},
	\langle R\cup R',\;\mathcal{Y}^{\star}\rangle
	\bigr),
\end{equation}
where $R \cup R'$ provides comprehensive case-specific reasoning, and $\mathcal{Y}^{\star}$ imposes stylistic conformity and domain-appropriate rhetoric. The summary prompt $p_{\text{summary}}$ guides the LLM to coherently integrate case data with retrieved expert exemplars(\textcolor{black}{see  Supplementary Fig. 6}). This iterative refinement and retrieval structure markedly improves the factual completeness, internal consistency, and legal robustness of the output.

To further improve forensic trust and accountability, we embed this architecture within a human-in-the-loop (HITL) framework~\cite{hitl}. Following initial analysis generation, forensic or medical experts are invited to review, critique, and amend the generated text. Their interventions---ranging from logical corrections to rhetorical adjustments---serve as implicit reward signals guiding subsequent system outputs toward expert-aligned reasoning and interpretative preferences. 
The Collective Summarization Agent captures this feedback and synthesizes it into an updated output version, maintaining a transparent revision trail. Upon expert approval, the finalized analysis is passed to a downstream Forensic-LLM for definitive cause-of-death determination.
Forensic-LLM is a domain-adapted large language model based on the \texttt{DeepSeek-R1-Distill-Llama-8B}~\cite{deepseek-r1} architecture and fine-tuned using LoRA, which efficiently adapts large models by introducing trainable low-rank matrices into frozen pre-trained weights. Specifically, for a weight matrix $W \in \mathbb{R}^{d \times k}$, LoRA re-parameterizes it as:
\begin{equation}
	W' = W + \Delta W = W + A B,
\end{equation}
where $A \in \mathbb{R}^{d \times r}$ and $B \in \mathbb{R}^{r \times k}$ with $r \ll \min(d, k)$. In our implementation, LoRA is applied to all linear layers ($\texttt{lora\_target: all}$) with a rank $r=8$, utilizing \texttt{bf16} precision, a learning rate of $1 \times 10^{-4}$, cosine decay scheduling, and a warm-up ratio of 0.1. The fine-tuning dataset includes all $6,739$ corpus entries and their associated conclusions, enabling the model to internalize forensic reasoning structures and stylistic norms.
The complete system outputs:
\begin{itemize}
	\item \textbf{Long-Form Analysis}: A detailed narrative capturing causal logic, evidence integration, and interpretative depth.
	\item \textbf{Short-Form Conclusion}: A succinct, decision-ready summary of the cause-of-death suitable for reporting and archival.
\end{itemize}

\subsubsection*{Reflection \& Memory Module}
The Reflection \& Memory Module maintains a long-horizon, low-entropy knowledge base that continuously aggregates the intermediate findings of the Local Solver while ensuring global consistency across the entire multi-agent system. At the $t$-th reasoning step, let $O_t$ be the raw textual output of the Local Solver and $ S_{t-1}=\{\,s_1,\dots,s_{t-1}\} $ the memory state prior to the update. We suppress the redundancy and hallucination information in the result section of the Local Solver through a reflective mechanism:
\begin{equation}
	R_t = \mathcal{M}(p_{\text{filter}},S_{t-1},O_t,B), 
\end{equation}
where $B$ denotes the forensic background content provided by the user, $\mathcal{M}$ denotes the LLM, and $p_{\text{filter}}$ represents the structured prompt(\textcolor{black}{see  Supplementary Fig. 7}).The filtering criterion theoretically is:
\begin{equation}
	\forall r\in R_t:\quad
	\bigl[\neg\ \exists s\in M_{t-1}\text{ s.t. }r\subseteq s\bigr]
	\;\land\;
	\bigl[r\text{ is consistent with }B\bigr].
\end{equation}
That is, every retained sentence $r$ must (i) contribute previously unseen semantic content and (ii) avoid contradiction with the established forensic background content.The pruned set $R_t$ is converted into a concise summary:
\begin{equation}
\tilde{s}_t \;=\; \mathcal{M}(p_{\text{tidy}},\,R_t,\,B),
\end{equation}
where $\tilde{s}_t$ is length-bounded to at most $k$ words to prevent memory bloat(\textcolor{black}{see  Supplementary Fig. 8}).The new memory state is obtained via $M_t \;=\; M_{t-1}\cup\{\tilde{s}_t\}.$
By iterating the mapping
\(
(O_t,M_{t-1})\mapsto M_t
\)
,this yields a compact yet comprehensive forensic memory that suppresses cumulative hallucinations, eliminates redundant information, and provides a consistent contextual substrate for the Planner, Local Solver, and Global Solver alike.

\subsubsection*{Backbone LLMs}
In this study, we orchestrate a heterogeneous model ensemble comprising two cloud-based LLMs, a unified vector-embedding backbone, and two on-premise LLMs to underpin our forensic-analysis framework.
The online models, i.e., DeepSeek-V3 (DeepSeek)~\cite{deepseek-v3} and GPT-4O-mini (OpenAI)~\cite{gpt4o}, provide high-capacity reasoning and planning capabilities, while the text-embedding-3-large model supplies consistent semantic representations for the RAG pipeline. Locally, we deploy Baichuan-M1~\cite{baichuan} (14 B parameters) and the domain-specific Forensic LLM (8 B parameters). DeepSeek-V3 drives every stage of the Planner Agent, from constructing diversified reasoning trajectories to node selection and adaptation. Within the Local Solver, DeepSeek-V3 serves as the routing oracle and powers the ReAct-style Tool Agent responsible for dynamic tool invocation; the Executor Agent likewise inherits DeepSeek-V3 as its core reasoning engine. The Global Solver, which performs collaborative summarization and inference synthesis, also leverages DeepSeek-V3. To accelerate iterative interaction and optimize memory consolidation, the Reflection \& Memory module employs GPT-4O-mini. For deterministic behaviour and reproducible outputs, all cloud-based models operate with a temperature setting of 0. Furthermore, DeepSeek-V3 can be interchanged with peer state-of-the-art models such as GPT-4O~\cite{gpt4o} or Claude 3.5~\cite{claude} without architectural modifications, offering pragmatic flexibility in response to computational constraints or deployment preferences.

\subsection*{Data Collection and Curation}

For data construction, we systematically assembled a large-scale medicolegal corpus comprising $7,748$ death investigations drawn from six leading Chinese institutions: Shaanxi Baimei Forensic Judicial Appraisal Institute ($739$ cases), Xi‘an Jiaotong University Forensic Identification Center ($864$ cases), Sun Yat-sen University Forensic Identification Center ($2,851$ cases), Hebei Medical University Forensic Center ($1,847$ cases), Xinxiang Medical University Forensic Identification Center ($735$ cases), and Jining Medical University Forensic Identification Center ($712$ cases).
Each record preserves multimodal evidence streams: decedent demographics, detailed case context, pre-mortem clinical findings, complete autopsy results (gross and histopathological), toxicology screens, long-form expert analyses, and short-form conclusions codifying the final cause-of-death determination. 
Guided by the national forensic pathology textbook~\cite{Forensic pathology,Knight's} and consensus guidelines, we annotated $15$ prevalent death aetiologies---Burning, Coronary Artery Atherosclerosis, Cerebral Hemorrhage, Drowning, Electrical Injury, High-Fall Trauma, Mechanical Asphyxia, Myocarditis, Peritonitis, Pneumonia, Poisoning, Traumatic Abdominal Injury, Traumatic Brain Injury, Traumatic Lung Injury, and deaths With Inducing Cause.
A two-stage triage pipeline was then employed: an AI-based classifier first filtered cases for completeness and diagnostic clarity, after which senior forensic pathologists performed targeted adjudication. This human-in-the-loop procedure retained $5,068$ cases with fully coherent multimodal documentation and unambiguous causal labels, while $2,680$ less-comprehensive or infrequent-class cases were consolidated as ``Other''.
The verified and cause-of-death categorised cases were randomly stratified into training ($4,059$; $80 \%$) and test ($1,009$; $20 \%$) partitions, preserving class balance to facilitate development and evaluation of the FEAT intelligent-agent framework.
Finally, the ``Other'' cohort was merged with the training subset, yielding a $6,739$-record corpus that underpinned two downstream tasks: instruction-tuning a domain-specific Forensic LLM by mapping long-form expert analyses to their corresponding short-form conclusions, and constructing the H-RAG index in which each long-form expert analysis serves as a dense-vector entry capturing structured forensic reasoning for efficient evidence-aware retrieval.

\subsection*{Comparisons with Representative AI Models} 

To rigorously evaluate the performance of \textsc{FEAT}, we conducted a comparative analysis with three representative baseline models, including the state-of-the-art AI agent systems, specifically MedAgent~\cite{medagent}, adapted for forensic pathology, and two powerful general-purpose LLMs, i.e., GPT-4O~\cite{gpt4o} and Claude3.5-Sonnet~\cite{claude}, guided by detailed CoT prompting~\cite{cot}.

The first comparative model, \textbf{MedAgent}, implements a training-free, multi-agent role-playing pipeline. Originally designed for medical inquiries, MedAgent begins by assembling virtual domain experts who independently draft analyses, subsequently synthesizing these into an integrated report refined through iterative debate until consensus is reached. For an equitable and domain-relevant comparison, we adapted MedAgent for forensic pathology applications by modifying its initial step: multiple virtual forensic pathology experts were tasked with independently drafting analyses from diverse perspectives. All other steps of MedAgent remained consistent with its original design. Notably, MedAgent employed the same backbone LLMs and hyperparameter settings as \textsc{FEAT}, with three methodologically key distinctions: 
its reasoning pipeline employs a noticeably shallower planning horizon, restricting the depth of evidentiary exploration that is crucial for robust forensic inference; it omits any form of tool-augmented retrieval or calculation, thereby foregoing the ability to incorporate external domain-specific resources at inference time; and it depends solely on remote, non-specialized LLM endpoints rather than a locally fine-tuned Forensic LLM, a choice that not only introduces variability in latency and consistency but also undermines the agent's capacity to distill a comprehensive Long-Form Analysis into a concise, forensic-domain defensible Short-Form Conclusion.

The remaining two comparative models, \textbf{GPT-4O-CoT} and \textbf{Claude3.5-Sonnet-CoT}, represent leading general-purpose LLMs renowned for their robust performance across diverse tasks. For these models, we meticulously designed prompts adhering to a detailed chain-of-thought methodology. Each prompt encompassed a comprehensive task overview, step-by-step instructions tailored explicitly to forensic pathology death analysis, illustrative examples, and clearly defined input content. These structured prompts, authored and validated by forensic pathology experts, guided the LLMs to systematically generate forensic analyses and subsequently derive definitive cause-of-death conclusions. For instance, prompts included scenario-specific instructions such as, ``Given the autopsy findings and toxicology report, explain the most likely pathophysiological sequence leading to death,'' accompanied by sample answers to demonstrate reasoning depth and expected format. This rigorous prompting approach facilitated direct evaluation of GPT-4O and Claude3.5-Sonnet's analytical capabilities against \textsc{FEAT}.

\subsection*{Human (Expert) Evaluations}

In this study, we conducted a rigorous human evaluation involving four renowned forensic pathology experts from prestigious institutions in China: Fudan University, Huazhong University of Science and Technology, Shanghai Forensic Science Institute, and Hebei Medical University. Each expert possessed over $20$ years of extensive experience in forensic pathology, making them among the top specialists in the field nationally.
Our evaluation utilized two distinct sets of forensic cases, each comprising $125$ unique and non-overlapping cases obtained from the test dataset through stratified random sampling across cause-of-death categories, thereby preventing potential biases and ensuring robust comparability between evaluations.
For each case, experts addressed eight specific evaluation questions designed to rigorously assess the performance and reliability of our FEAT system.

The first type of question presented the experts with three sets of outputs for comparison: results generated solely by FEAT (without human-in-the-loop), results produced through interactions between FEAT and doctoral-level forensic students (with human-in-the-loop), and the original expert forensic analyses, including the detailed reasoning processes and conclusions. The order of the answers to each test is randomly upset. Experts evaluated the outputs based on the following criteria:
\begin{enumerate}[i.]
	\item \textbf{Better reflects forensic consensus}: Which output best reflects the current scientific consensus in forensic science and pathology, including accepted standards for cause-of-death analysis and autopsy procedures?
	\item \textbf{Better forensic question comprehension}: Which output demonstrates a superior understanding of forensic pathology questions, accurately employing forensic terminology and adhering to relevant legal definitions?	
	\item \textbf{Better forensic knowledge recall}: Which output more effectively recalls and integrates relevant forensic pathology knowledge, such as autopsy findings, cause-of-death analyses, and toxicological reports?	
	\item \textbf{Better forensic reasoning}: Which output better demonstrates the ability to logically apply forensic pathology knowledge or evidence analysis to infer the cause and manner of death?	
	\item \textbf{More inaccurate/irrelevant information}: Which output includes more irrelevant or inaccurate content that may lead to misunderstandings or incorrect conclusions regarding forensic evidence?	
	\item \textbf{Omits more key information}: Which output omits more critical forensic pathology information, such as key autopsy findings, toxicology results, or pertinent legal considerations that could affect case outcomes?	
	\item \textbf{More evidence of demographic bias}: Which output contains information potentially biased against particular demographic groups based on gender, race, age, or other factors, thereby affecting the impartiality of forensic evidence interpretation? 	
	\item \textbf{Greater potential for harm}: Which output poses a greater risk of harm due to inaccuracies or misinterpretations in the autopsy report, potentially adversely affecting criminal investigations, judicial outcomes, or public safety?
\end{enumerate}
Experts were permitted to select one or multiple outputs per criterion, allowing precise quantification of performance gaps between FEAT-generated results and established forensic standards.

In the second type of evaluation, experts assessed outputs generated solely by FEAT (without human-in-the-loop) using a three-tiered rating system---fully acceptable, partially acceptable, and unacceptable---to clearly distinguish the degrees of model accuracy, applicability, and adherence to forensic standards. The assessment criteria were:
\begin{enumerate}[i.]
	\item \textbf{Better forensic-case understanding}: Does the model demonstrate comprehensive and accurate comprehension of forensic cases, including autopsy reports, medical histories, and relevant evidence?  
	
	\item \textbf{Better adherence to forensic standards}: Does the model-generated output adhere to established forensic pathology guidelines, legal standards, and relevant protocols?
	
	\item \textbf{Better forensic reasoning}: Is the model capable of employing logical reasoning effectively to correlate forensic evidence and accurately determine the cause and manner of death?
	
	\item \textbf{Better forensic differential diagnosis}: Can the model accurately distinguish among multiple potential causes of death (e.g., natural, accidental, suicide, homicide) based on forensic evidence? 
	
	\item \textbf{Better reliability/direct usability of forensic findings}: Is the model-generated forensic output reliable, scientifically valid, and suitable for legal proceedings or further investigations?
	
	\item \textbf{More inaccurate/irrelevant forensic-report content}: Does the model-generated forensic report avoid inaccuracies, misleading statements, or false information?  
	
	\item \textbf{More bias in forensic analysis}: Does the model maintain objectivity and fairness, ensuring its forensic analyses are free from biases related to age, gender, culture, or ethnicity?
	
	\item \textbf{Greater potential forensic harm}: Could the model-generated forensic conclusions potentially result in adverse consequences such as wrongful judgments, evidence misinterpretation, or other negative impacts?
\end{enumerate}

This structured and comprehensive evaluation ensured robust scrutiny of FEAT's outputs, emphasizing the validity, reliability, and potential real-world applicability of the model in forensic investigations. These findings have the potential to directly inform forensic practices and policies, enhancing the quality of forensic analyses and supporting more accurate and fair judicial outcomes.

\subsection*{Statistical Analyses}

All statistical procedures were executed in \texttt{Python}~3.11 (Python Software Foundation) with the scientific-computing ecosystem \texttt{NumPy}~(v1.24.4), \texttt{pandas}~(v2.2.3) and \texttt{SciPy}~(v1.13.0).
For human(experts)-evaluation experiments, two-sided 95\% confidence intervals (CIs) for class-level proportions were estimated by non-parametric bootstrap resampling. Specifically, $10^{5}$ bootstrap samples were drawn with replacement, and the normal approximation to the bootstrap sampling distribution was used to derive symmetric CIs, a resample size chosen empirically to stabilise the interval width ($<0.1$ percentage-point change upon further doubling of resamples).  
To compare model performance across competing methodological variants, we adopted a large-language-model-based semantic-similarity metric. Both agent(model)-generated answers and expert-annotated references were embedded with \texttt{text-embedding-3-large} (OpenAI), and pairwise cosine similarities were computed. The resulting \textbf{OpenAI-score} serves as a quantitative proxy for conceptual concordance, with higher scores indicating closer alignment to expert reasoning.

\subsection*{Data Availability} 

Due to privacy protection, we will gradually open up our data base for use by a wider range of scholars.

\subsection*{Code Availability} 

The code will be available at \href{https://github.com/shenxiaochenn/FEAT}{https://github.com/shenxiaochenn/FEAT}.

\def\bibsection{\section*{\refname}}

\end{document}